\documentclass{article} 

\usepackage{iclr2017_conference,times}

\usepackage[pagebackref=true,breaklinks=true,letterpaper=true,colorlinks,bookmarks=false,citecolor=blue]{hyperref}
\usepackage{url}
\usepackage{import}
\usepackage[T1]{fontenc}    
\usepackage{url}            
\usepackage{booktabs}       
\usepackage{amsfonts}       
\usepackage{nicefrac}       

\usepackage{graphicx}
\usepackage{soul}
\usepackage{color}
\usepackage{epsfig}
\usepackage[belowskip=0pt,aboveskip=0pt,font=small]{caption}
\usepackage[belowskip=0pt,aboveskip=0pt,font=small]{subcaption}
\usepackage{amsmath, amsthm, amssymb}
\usepackage{textcomp}
\usepackage{stmaryrd}
\usepackage{upgreek}
\usepackage{bm}
\usepackage{subcaption}
\usepackage{cases}
\usepackage{mathtools}
\usepackage{multirow}
\usepackage{rotating}
\usepackage{enumitem}
\usepackage[olditem,oldenum]{paralist}
\usepackage{alltt}
\usepackage{listings}
\usepackage{mysymbols}
\usepackage{bm}
\usepackage{xspace}
\usepackage{multirow}
\usepackage{comment}
\usepackage{afterpage}
\usepackage{pdfpages}
\usepackage{framed}
\usepackage{fancybox}
\usepackage{epstopdf}
\usepackage[ruled,vlined,oldcommands,linesnumbered]{algorithm2e}
\SetCommentSty{mycommfont}
\usepackage{tikz}
\usetikzlibrary{arrows,shapes, calc, shapes.misc}

\newcommand{\xhdr}[1]{{\vspace{2pt}\noindent\textbf{#1}}}
\newcommand{\ar}[1]{\textcolor{black}{#1}} 

\title{Diverse beam Search: \\ Decoding Diverse Solutions from \\ Neural Sequence Models}

\author{Ashwin K Vijayakumar$^1$, Michael Cogswell$^1$, Ramprasath R. Selvaraju$^1$, Qing Sun$^1$ \\
\textbf{Stefan Lee$^1$, David Crandall$^2$ \& Dhruv Batra$^{1}$} \\
\texttt{\{ashwinkv,cogswell,ram21,sunqing,steflee\}@vt.edu} \\
\texttt{djcran@indiana.edu},
\texttt{dbatra@vt.edu} \\
\\
$^1$ Department of Electrical and Computer Engineering,  \\
Virginia Tech, Blacksburg, VA, USA \\
\\
$^2$ School of Informatics and Computing\\
Indiana University, Bloomington, IN, USA
}

\begin{document}

\maketitle
\begin{abstract}
Neural sequence models are widely used to model time-series data. 
Equally ubiquitous is the usage of beam search (BS) as an approximate inference algorithm to decode output sequences from these models. 
BS explores the search space in a greedy left-right fashion retaining only the top-$B$ candidates -- resulting in sequences that differ only slightly from each other. 
Producing lists of nearly identical sequences is not only computationally wasteful but also typically fails to capture the inherent ambiguity of complex AI tasks.
To overcome this problem, we propose \emph{Diverse Beam Search} (DBS), an alternative to BS that
decodes a list of diverse outputs by optimizing for a diversity-augmented objective.
We observe that our method finds better top-1 solutions by controlling for the exploration and exploitation of the search space -- implying that DBS is a \emph{better search algorithm}.
Moreover, these gains are achieved with minimal computational or memory overhead as compared to beam search.
To demonstrate the broad applicability of our method, we present results on image captioning, machine translation and visual question generation using both standard quantitative metrics and qualitative human studies. 
\ar{Further, we study the role of diversity for image-grounded language generation tasks as the complexity of the image changes.}
\ar{We observe that} our method consistently outperforms BS and previously proposed techniques for diverse decoding from neural sequence models.
\end{abstract}

\section{Introduction} \label{intro}
\newcommand{\yb}{\mathbf{y}}
\newcommand{\xb}{\mathbf{x}}

In the last few years, Recurrent Neural Networks (RNNs), Long Short-Term Memory networks (LSTMs) or more generally, neural sequence models have become the standard choice for modeling time-series data for a wide range of applications such as speech recognition \citep{graves_arxiv13},  machine translation \citep{bahdanau_arxiv14}, conversation modeling \citep{vinyals_arxiv15}, image and video captioning \citep{vinyals_cvpr15, venugopalan_cvpr15}, and visual question answering \citep{antol_iccv15}. 
RNN based sequence generation architectures model the conditional probability, $\prob(\yb | \xb)$ of an output sequence $\yb = (y_1,\ldots,y_T)$ given an input $\xb$ (possibly also a sequence); where the output tokens $y_t$ are from a finite vocabulary, $\calV$.

\begin{figure}[t]
\begin{center}
\includegraphics[width=1\columnwidth]{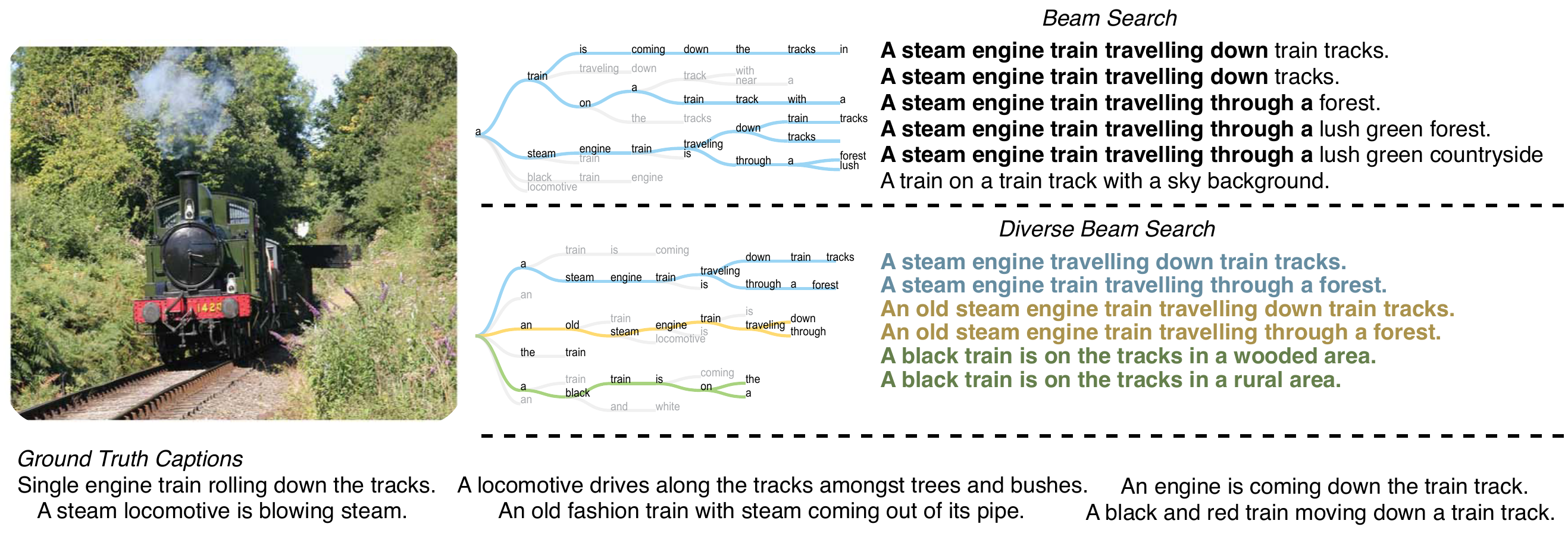}
\vspace{0.5pt}
\caption{
Comparing image captioning outputs decoded by BS and our method, Diverse Beam Search (DBS) -- we notice that BS captions are near-duplicates with similar shared paths in the search tree and minor variations in the end. In contrast, DBS captions are significantly diverse and similar to the inter-human variability in describing images.}
\label{fig:cover}
\end{center}
\end{figure}

\textbf{Inference in RNNs.} Maximum a Posteriori (MAP) inference for RNNs is te task of finding the most likely output sequence given the input. Since the number of possible sequences grows as $|\calV|^T$, exact inference is NP-hard so approximate inference algorithms like Beam Search (BS) are commonly employed. 
BS is a heuristic graph-search algorithm that maintains the $B$ top-scoring partial sequences expanded in a greedy left-to-right fashion. Fig.~\ref{fig:cover} shows a sample BS search tree.

\textbf{Lack of Diversity in BS.} Despite the widespread usage of BS, it has long been understood that solutions decoded by BS are generic and lacking in diversity~\citep{finkel_emnlp06,gimpel_emnlp13,li_arxiv15,li_arxiv16}.
To illustrate this, a comparison of captions provided by humans (bottom) and BS (topmost) are shown in \figref{fig:cover}. 
While this behavior of BS is disadvantageous for many reasons, we highlight the three most crucial ones here:
\setlength{\plitemsep}{-1.5ex}
\begin{compactenum}[\hspace{-1pt}i)]
\item The production of near-identical beams make BS a computationally wasteful algorithm, with essentially the same computation being repeated for no significant gain in performance.\\
\item Due to \emph{loss-evaluation mismatch} \ie improvements in posterior-probabilities not necessarily corresponding to improvements in task-specific metrics, it is common practice \citep{vinyals_cvpr15, karpathy_cvpr15, ferraro_arxiv16} to \emph{deliberately throttle BS to become a poorer optimization algorithm} by using reduced beam widths.
This treatment of an optimization algorithm as a hyper-parameter is not only intellectually dissatisfying but also has a significant practical side-effect -- it leads to the decoding of largely bland, generic, and ``safe'' outputs, \eg always saying ``I don't know'' in conversation models~\citep{corrado_blog15}. \\
\item Most importantly, lack of diversity in the decoded solutions is fundamentally crippling in AI problems with \emph{significant ambiguity} -- \eg there are multiple ways of describing an image or responding in a conversation that are ``correct'' and it is important to capture this ambiguity by finding several diverse plausible hypotheses. 
\end{compactenum}	
\setlength{\plitemsep}{1ex}

\textbf{Overview and Contributions.} To address these shortcomings, we propose \emph{Diverse Beam Search (DBS)} -- a general framework to decode a list of diverse sequences that can be used as an \emph{alternative} to BS. 
At a high level, DBS decodes diverse lists by dividing the beam budget into groups and enforcing diversity between groups of beams. 
Drawing from recent work in the probabilistic graphical models literature on Diverse M-Best (\texttt{DivMBest}) MAP inference~\citep{batra_eccv12, prasad_nips14,kirillov_cvpr15}, we optimize an objective that consists of two terms -- the sequence likelihood under the model and a dissimilarity term that encourages beams across groups to differ. 
This diversity-augmented model score is optimized in a \emph{doubly greedy} manner -- greedily optimizing along both time (like BS) and groups (like DivMBest).
\\ \\ 
To summarize, our primary technical contribution is Diverse Beam Search, a doubly greedy approximate inference algorithm for decoding diverse sequences.
\ar{To demonstrate its broad applicability, we report results on two image-grounded language generation tasks, captioning and question generation and on machine translation.
Our method consistently outperforms BS while being comparable in terms of both run-time and memory requirements.  
We find that DBS results in improvements on both oracle task-specific and diversity-related metrics against baselines. 
Further, we notice that these gains are more pronounced as the image becomes more complex consisting of multiple objects and interactions.}
We \ar{also} conduct human studies to evaluate the role of diversity in human preferences between BS and DBS for image captions.  We also analyze the parameters of DBS and show they are robust over a wide range of values. 
Finally, we also show that our method is general enough to incorporate various forms for the dissimilarity term. 
Our implementation is available at \url{https://github.com/ashwinkalyan/dbs}. Also, a demo of DBS on image-captioning is available at \url{dbs.cloudcv.org}.

\section{Preliminaries: Decoding RNNs with Beam Search}
\label{sec:prelim}
We begin with a refresher on BS, before describing our \ar{extension}, Diverse Beam Search. For notational convenience, let $[n]$ denote the set of natural numbers from $1$ to $n$ and let $\mathbf{v}_{[n]} = [v_1, v_2, \dots v_n]$ index the first $n$ elements of a vector $\mathbf{v} \in \RR^m$, where $n\leq m$.

\textbf{The Decoding Problem.} RNNs are trained to estimate the likelihood of sequences of tokens from a finite dictionary $\calV$ given an input $\xb$. 
The RNN updates its internal state and estimates the conditional probability distribution over the next output given the input and all previous output tokens. 
We denote the logarithm of this conditional probability distribution over all tokens at time $t$ as $\theta(y_t) = \log \prob(y_t | y_{t-1},\ldots,y_1, \xb)$. 
To simplify notation, we index $\theta(\cdot)$ with a single variable $y_t$; but it should be clear that it depends on the previous outputs, $\yb_{[t-1]}$ from the context.
The $\log$-probability of a partial solution (\ie the sum of $\log$-probabilities of all previous tokens decoded) can now be written as $\Theta(\yb_{[t]})=\sum_{\tau\in[t]} \theta(y_\tau)$.
The decoding problem is then the task of finding a sequence $\yb$ that maximizes $\Theta(\yb)$.

As each output is conditioned on all the previous outputs, decoding the optimal length-$T$ sequence in this setting can be viewed as MAP inference on $T$-order Markov chain with the $T$ nodes corresponding to output tokens.
Not only does the size of the largest factor in such a graph grow as $|\calV|^T$, but also requires wastefully forwarding of the RNN repeatedly to compute entries in the factors.
Thus, approximate algorithms are employed.

\textbf{Beam Search.} The most prevalent method for approximate decoding is BS, which stores the top-$B$ highly scoring candidates at each time step; where $B$ is known as the \emph{beam width}.
Let us denote the set of $B$ solutions held by BS at the start of time $t$ as $Y_{[t-1]} = \{\yb_{1,[t-1]},\dots,\yb_{B,[t-1]}\}$. 
At each time step, BS considers all possible single token extensions of these beams given by the set $\calY_t = Y_{[t-1]}\times\calV$ and selects the $B$ most likely extensions. More formally, at each step,
\begin{equation}
\label{eq: bs}
 Y_{[t]} \ \ = \argmax_{\yb_{1,[t]},\dots,\yb_{B,[t]} \in \calY_t} \sum_{b \in [B]} \Theta(\yb_{b,[t]}) ~~~~~~s.t.~~\yb_{i,[t]} \neq \yb_{j,[t]}
\end{equation}
%
The above objective can be trivially maximized by sorting all $B\times|\calV|$ members of $\calY_t$ by their $\log$-probabilities and selecting the top-$B$.
This process is repeated until time $T$ and the most likely sequence is selected by ranking the B beams based on $\log$-probabilities. 

While this method allows for multiple sequences to be explored in parallel, most completions tend to stem from a single highly valued beam -- resulting in outputs that are typically only minor perturbations of a single sequence. 

\section{Diverse Beam Search: Formulation and Algorithm} 
\label{form}

To overcome this shortcoming, we consider augmenting the objective in Eq. \ref{eq: bs} with a dissimilarity term $\Delta(Y_{[t]})$ that measures the diversity between candidate sequences. Jointly optimizing for all $B$ candidates at each time step is intractable as the number of possible solutions grows with $|\calV|^B$ (which can easily reach $10^{60}$ for typical language modeling settings).
To avoid this joint optimization problem, we divide the beam budget $B$ into $G$ groups and greedily optimize each group using beam search while holding previous groups fixed. 
This doubly greedy approximation along both time and across groups turns $\Delta(Y_{[t]})$ into a function of only the current group's possible extensions. We detail the specifics of our approach in this section.

\textbf{Diverse Beam Search.} Let $Y_{[t]}$, the set of all $B$ beams at time $t$ be partitioned into $G$ {non-empty}, {disjoint} subsets $Y_{[t]}^g, \ g{\in}[G]$. 
Without loss of generality, consider an equal partition such that each group contains $B'=\nicefrac{B}{G}$ groups. 
Beam search can be applied to each group to produce $B$ solutions; however, each group would produce identical outputs.

\ar{Unlike BS, we optimize a modified version of the objective of eq. \ref{eq: bs} which adds a dissimilarity term $\Delta(\yb_{[t]}, Y_{[t]}^g)$, measuring the dissimilarity of a sequence $\yb_{[t]}$ against a group $Y_{[t]}^g$.}
While $\Delta(\cdot, \cdot)$ can take various forms, for simplicity we define one broad class that decomposes across beams within each group as:
\begin{equation}\label{eq:diversity-term}
    \Delta(\yb_{[t]}, Y_{[t]}^g) = \sum_{b=1}^{B'}\delta\left(\yb_{[t]}, \yb_{b,[t]}^g\right)
\end{equation}
where $\delta(\cdot, \cdot)$ is a measure of sequence dissimilarity -- \eg a negative cost for each co-occurring n-gram in two sentences or distance between distributed sentence representations.
The exact form of the sequence-level dissimilarity term can vary and we discuss some choices in \secref{div_type}.

As we optimize each group with the previous groups fixed, extending group $g$ at time $t$ amounts to a standard BS using dissimilarity augmented $\log$-probabilities and can be written as: 
\begin{eqnarray}
\label{eq:divminB}
Y^g_{[t]} \ \  = \argmax_{\yb^g_{1,[t]}, \dots, \yb^g_{B',[t]} \in \mathcal{Y}^g_{t}} &&  \sum_{b \in [B']} \Theta(\yb^g_{b,[t]}) + \sum_{h=1}^{g-1}\lambda_g\Delta\left(\yb_{b,[t]}^g, Y_{[t]}^h\right)\\
&& \st \ \yb^g_{i,[t]} \neq \yb^g_{j,[t]},\  \lambda_g \geq 0 \nonumber
\end{eqnarray}

\tikzstyle{g1vertex}=[rounded rectangle,fill=green!25,minimum size=20pt,inner sep=3pt]
\tikzstyle{g2vertex}=[rounded rectangle,fill=red!25,minimum size=20pt,inner sep=3pt]
\tikzstyle{g3vertex}=[rounded rectangle,fill=blue!25,minimum size=20pt,inner sep=3pt]
\tikzstyle{g3shade}=[rounded rectangle,opacity=0.65, fill=blue!25,minimum size=20pt,inner sep=3pt]
\tikzstyle{edge} = [draw,thick,<-]
\tikzstyle{bedge} = [draw,opacity=0.75,thick,<-,bend right=35]
\tikzstyle{weight} = [font=\small]
\tikzstyle{selected edge} = [draw,line width=5pt,-,red!50]
\tikzstyle{ignored edge} = [draw,line width=5pt,-,black!20]
\tikzstyle{group box} = [thick, rounded corners=0.25cm, minimum width=30pt, minimum height=50pt]
\tikzstyle{searchbox} = [thick, fill=white, rounded corners=0.25cm, minimum width=120pt, minimum height=50pt]

\newcommand{\vertspace}{-1.15}
\newcommand{\inspace}{-0.45}
\newcommand{\of}{1.25}
\newcommand{\hz}{1.3}
\begin{figure}[t]
\centering
\resizebox{0.95\columnwidth}{!}{
\begin{tikzpicture}[scale=1.8, auto,swap]
	
		\node[inner sep=0pt] (input) at (-1.1,\vertspace-0.15){\includegraphics[width=.35\textwidth]{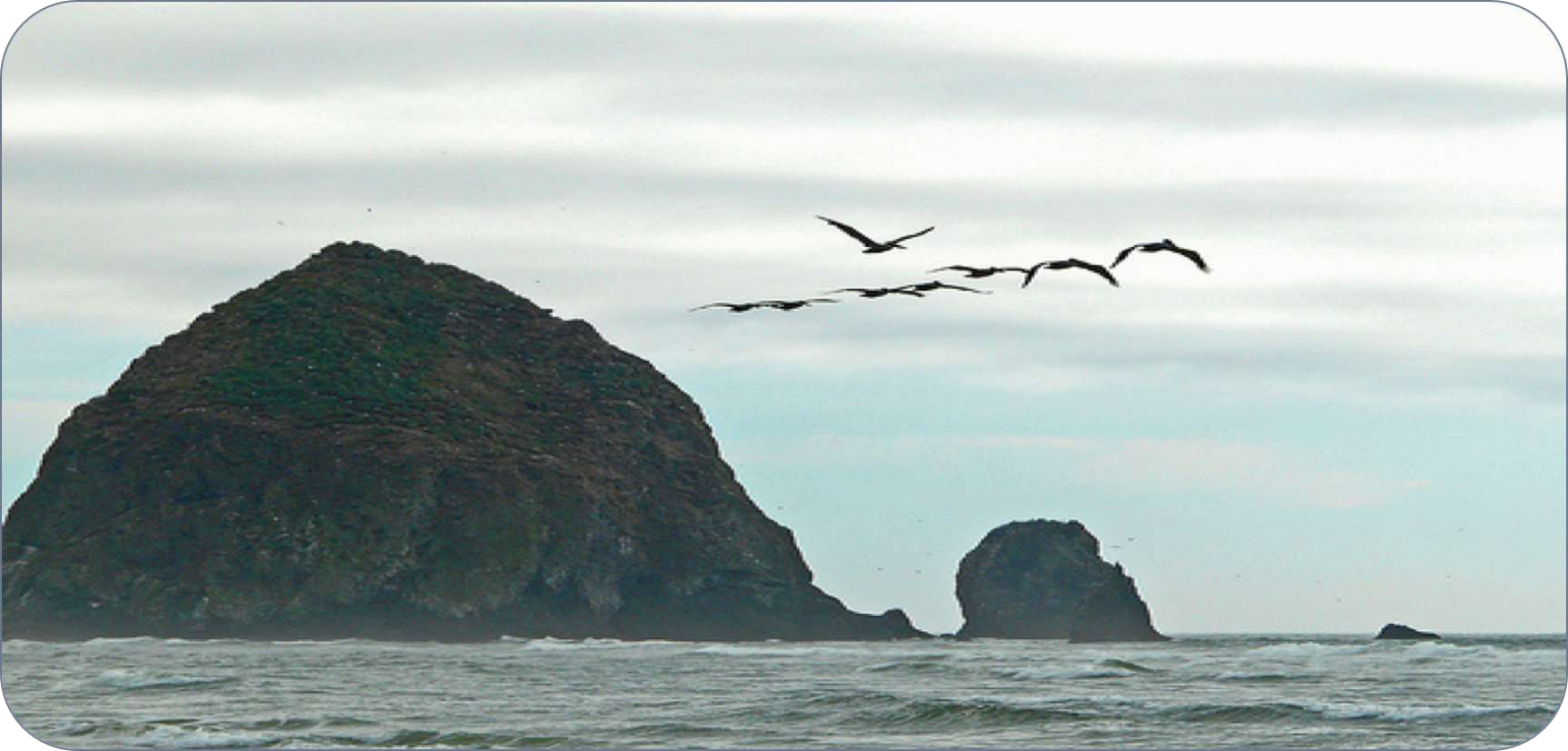}};
    
		\node[rotate=90](g1) at (\of-0.5,0.5*\inspace) {\textbf{Group 1}};
		\node[rotate=90](g2) at (\of-0.5,0.5*\inspace+\vertspace) {\textbf{Group 2}};
		\node[rotate=90](g3) at (\of-0.5,0.5*\inspace+2*\vertspace) {\textbf{Group 3}};

    \node[g1vertex](a11) at (0*\hz+\of,0) {a};
    \node[g1vertex](a21) at (1*\hz+\of,0) {flock};
    \node[g1vertex](a31) at (2*\hz+\of,0) {of};
    \node[g1vertex](a41) at (3*\hz+\of,0) {birds};
    \node[g1vertex](a51) at (4*\hz+\of,0) {flying};
    \node[g1vertex](a61) at (5*\hz+\of,0) {over};
    \node[g1vertex](b11) at (0*\hz+\of,\inspace) {a};
    \node[g1vertex](b21) at (1*\hz+\of,\inspace) {flock};
    \node[g1vertex](b31) at (2*\hz+\of,\inspace) {of};
    \node[g1vertex](b41) at (3*\hz+\of,\inspace) {birds};
    \node[g1vertex](b51) at (4*\hz+\of,\inspace) {flying};
    \node[g1vertex](b61) at (5*\hz+\of,\inspace) {in};
    \node[group box](11) at (0*\hz+\of,-0.22){};
    \node[group box](21) at (1*\hz+\of,-0.22){} edge[edge](11);
    \node[group box](31) at (2*\hz+\of,-0.22){} edge[edge](21);
    \node[group box](41) at (3*\hz+\of,-0.22){} edge[edge](31);
    \node[group box](51) at (4*\hz+\of,-0.22){} edge[edge](41);
    \node[group box](61) at (5*\hz+\of,-0.22){} edge[edge](51);

    \node[g2vertex](a12) at (0*\hz+\of,\vertspace) {birds};
    \node[g2vertex](a22) at (1*\hz+\of,\vertspace) {flying};
    \node[g2vertex](a32) at (2*\hz+\of,\vertspace) {over};
    \node[g2vertex](a42) at (3*\hz+\of,\vertspace) {the};
    \node[g2vertex](a52) at (4*\hz+\of,\vertspace) {water};
    \node[g2vertex](b12) at (0*\hz+\of,\vertspace+\inspace) {birds};
    \node[g2vertex](b22) at (1*\hz+\of,\vertspace+\inspace) {flying};
    \node[g2vertex](b32) at (2*\hz+\of,\vertspace+\inspace) {over};
    \node[g2vertex](b42) at (3*\hz+\of,\vertspace+\inspace) {an};
    \node[g2vertex](b52) at (4*\hz+\of,\vertspace+\inspace) {ocean};
    \node[group box](12) at (0*\hz+\of,-0.22+\vertspace){} edge[edge](11);
    \node[group box](22) at (1*\hz+\of,-0.22+\vertspace){} edge[edge](21) edge[edge](12);
    \node[group box](32) at (2*\hz+\of,-0.22+\vertspace){} edge[edge](31) edge[edge](22);
    \node[group box](42) at (3*\hz+\of,-0.22+\vertspace){} edge[edge](41) edge[edge](32);
    \node[group box](52) at (4*\hz+\of,-0.22+\vertspace){} edge[edge](51) edge[edge](42);

    \node[g3vertex](a13) at (0*\hz+\of,2*\vertspace) {several};
    \node[g3vertex](a23) at (1*\hz+\of,2*\vertspace) {birds};
    \node[g3vertex](a33) at (2*\hz+\of,2*\vertspace) {are};
    \node[g3vertex](b13) at (0*\hz+\of,2*\vertspace+\inspace) {several};
    \node[g3vertex](b23) at (1*\hz+\of,2*\vertspace+\inspace) {birds};
    \node[g3vertex](b33) at (2*\hz+\of,2*\vertspace+\inspace) {fly};
    \node[group box](13) at (0*\hz+\of,-0.22+2*\vertspace){} edge[bedge](11) edge[edge](12);
    \node[group box](23) at (1*\hz+\of,-0.22+2*\vertspace){} edge[bedge](21) edge[edge](22) edge[edge](13);
    \node[group box](33) at (2*\hz+\of,-0.22+2*\vertspace){} edge[bedge](31) edge[edge](32) edge[edge](23);
    \node[group box](43) at (3*\hz+\of,-0.22+2*\vertspace){} edge[bedge](41) edge[edge](42);
    
    \node[draw,searchbox, inner sep=0pt](search) at (4.17*\hz+\of,-0.22+2*\vertspace){
    \hspace{18pt}
    \scriptsize
    \begin{tabular}{l}
    \textbf{~Modify scores to include diversity:}\\
    $\quad \theta(`the')+\lambda\Delta(`birds',`the', `an')[`the']$ \\[-5px]
    \multicolumn{1}{c}{\vdots} \\
    $\quad \theta(`over')+\lambda\Delta(`birds',`the', `an')[`over']$
    \end{tabular}
    } edge[edge](33);

    \node[g3shade](a43) at (3*\hz+\of,2*\vertspace) {?};
    \node[g3shade](b43) at (3*\hz+\of,2*\vertspace+\inspace) {?};

    \node[g1vertex](s1) at (7.4*\hz+\of, 0){a flock of birds flying over the ocean};
    \node[g1vertex](s2) at (7.4*\hz+\of, \inspace){a flock of birds flying over a beach};
    \node[g2vertex](s1) at (7.4*\hz+\of, \vertspace){birds flying over the water in the sun};
    \node[g2vertex](s2) at (7.4*\hz+\of, \vertspace+\inspace){birds flying the water near a mountain};
		\node[g3vertex](s1) at (7.4*\hz+\of, 2*\vertspace){several birds are flying over a body of water};
    \node[g3vertex](s2) at (7.4*\hz+\of, 2*\vertspace+\inspace){several birds flying over a body of water};

    \draw[color=black,thick, ->] (6.7*\hz, 0.4) -- node[auto, left, rotate=90, yshift=-10pt, xshift=20pt] {$\footnotesize groups$} (6.7*\hz, -2.3*\hz);

    \draw[color=black,thick, ->] (\of-0.25, 0.4) -- node[auto, above] {$\footnotesize t$} (5*\hz+\of+0.25, 0.4);

\end{tikzpicture}
}
\vspace{10pt}
\caption{Diverse beam search operates left-to-right through time and top to bottom through groups. Diversity between groups is combined with joint log-probabilities, allowing continuations to be found efficiently. The resulting outputs are more diverse than for standard approaches.}
\label{fig:overview}
\vspace{-15pt}
\end{figure}
This approach, which we call Diverse Beam Search (DBS) is detailed in Algorithm \ref{alg:dbs}. 
An example run of DBS is shown in Figure \ref{fig:overview} for decoding image-captions. 
In the example, $B{=}6$ and $G{=}3$ and so, each group performs a smaller, diversity-augmented BS of size 2. 
In the snapshot shown, group 3 is being stepped forward and the diversity augmented score of all words in the dictionary is computed conditioned on previous groups. 
The score of all words are adjusted by their similarity to previously chosen words -- `birds', `the' and `an' (Algorithm \ref{alg:dbs}, Line \ref{alg:aug}).
The optimal continuations are then found by standard BS (Algorithm \ref{alg:dbs}, Line \ref{alg:dobs}).

\begin{algorithm}[h]
\caption{Diverse Beam Search}
\label{alg:dbs}
Perform a diverse beam search with $G$ groups using a beam width of $B$ \\
\For{$t=1,\ \ldots \,T$}{
    \tcp{perform one step of beam search for first group without diversity}
    $Y_{[t]}^1 \leftarrow \argmax_{(\yb_{1,[t]}^1, \dots, \yb_{B',[t]}^1)} \sum_{b\in[B']}\Theta(\yb_{b,[t]}^1)$ \\
    \For{$g=2,\ \ldots \,G$}{
        \tcp{augment log-probabilities with diversity penalty}
    $\Theta(\yb_{b,[t]}^g) \leftarrow \Theta(\yb_{b,[t]}^g) + \sum_h\lambda_g \Delta(\yb_{b,[t]}^g, Y_{[t]}^h) \quad$ $b \in [B'], \yb_{b,[t]}^g \in \mathcal{Y}^g$ and $\lambda_g > 0$ \label{alg:aug}\\ 
        \tcp{perform one step of beam search for the group}
        $Y_{[t]}^g \leftarrow \argmax_{(\yb_{1,[t]}^g, \dots, \yb_{B',[t]}^g)} \sum_{b\in[B']}\Theta(\yb_{b,[t]}^g)$ \label{alg:dobs} \\
    }
}
Return set of B solutions, $Y_{[T]} = \bigcup^G_{g=1} Y_{[T]}^g$ 
\end{algorithm}

There are a number of advantages worth noting about our approach. 
By encouraging diversity between beams at each step (rather than just between highest ranked solutions like in \citet{gimpel_emnlp13}, our approach rewards each group for spending its beam budget to explore different parts of the output space rather than repeatedly chasing sub-optimal beams from prior groups. 
Furthermore, the staggered group structure enables each group beam search to be performed in parallel with a time offset. 
This parallel algorithm completes in $T + G$ time steps compared to $T\times G$ running time for a black-box approach of \citet{gimpel_emnlp13}. 
\\ \\
In summary, DBS is a task agnostic, doubly greedy algorithm that incorporates diversity in beam search with little memory or computational overhead.
Moreover, as the first group is not conditioned on other groups during optimization, our method is guaranteed to be at least as good as a beam search of size $B/G$.

\section{Related Work} \label{rel}

\xhdr{Diverse M-Best Lists.}
The task of generating diverse structured outputs from probabilistic models has been studied extensively \citep{park_iccv11,batra_eccv12,kirillov_cvpr15,prasad_nips14}.
\cite{batra_eccv12} formalized this task for Markov Random Fields as the \texttt{DivMBest} problem and presented a greedy approach which solves for outputs iteratively, conditioning on previous solutions to induce diversity. 
\cite{kirillov_cvpr15} show how these solutions can be found jointly for certain kinds of energy functions.
The techniques developed by Kirillov are not directly applicable to decoding from RNNs, which do not satisfy the assumptions made. 

Most related to our proposed approach is that of \cite{gimpel_emnlp13} who apply the \texttt{DivMBest} approach to machine translation using beam search as a black-box inference algorithm. 
To obtain diverse solutions, beam searches of arbitrary size are sequentially performed while retaining the top-scoring candidate and using it to update the diversity term.
This approach is extremely wasteful because in each iteration only one solution returned by beam search is kept. 
Consequently, the iterative method is time consuming and is poorly suited for batch processing or producing a large number of solutions. 
Our algorithm avoids these shortcomings by integrating diversity within BS such that \emph{no} beams are discarded. 
By running multiple beam searches \emph{in parallel} and at staggered time offsets, we obtain large time savings making our method comparable to classical BS. 
One potential advantage over our method is that more complex diversity measures at the sentence-level can be incorporated. 
However, as observed empirically by us and \cite{li_arxiv15}, initial words
tend to significantly impact the diversity of the resultant sequences -- suggesting that later words may not contribute significantly to diverse inference.

\xhdr{Diverse Decoding for RNNs.}
Some efforts have been made to produce diverse decodings from recurrent models for conversation modeling and machine translation. 

In this context, our work is closely related to \cite{li_arxiv16}, who propose a BS diversification heuristic to overcome the shortcomings of \cite{gimpel_emnlp13}. 
This discourages sequences from sharing common roots, implicitly resulting in diverse lists.
Introducing diversity through a modified objective as in DBS rather than a heuristic provides easier generalization to incorporate different notions of diversity and control for the exploration-exploitation trade-off as detailed in \secref{div_type}.
Furthermore, we find that DBS outperforms this method.

Through a novel decoding objective that maximizes mutual information between inputs and predicted outputs, \cite{li_arxiv15} penalize decoding generic, input independent sequences. 
This is achieved by training an additional target language model.
Although this work and DBS share the same goals (producing diverse decodings), the techniques developed are disjoint and complementary -- \cite{li_arxiv15} develops a new model (RNN translation model with an RNN target language model), while DBS is a modified \emph{inference} algorithm that can be applied to \emph{any} model where BS is applicable. 
Combination of these complementary techniques is left as interesting future work.

\vspace{-10pt}
\section{Experiments} \label{sec:results}
\vspace{-5pt}
We first explain the baselines and evaluation metrics used in this paper. 
Next, we proceed to the analysis of the effects of DBS parameters.
Further, we report results on image-captioning, machine translation and visual question generation.
\ar{In the context of image-grounded language generation tasks, we additionally study the role of diversity with varying \emph{complexity} of the image.}
Although results are reported on these tasks, it should be noted that DBS is a task-agnostic algorithm that can replace BS to decode diverse solutions.

\xhdr{Baselines.} We compare with beam search and the following existing methods: 
\begin{compactenum}[\hspace{0pt} -]
\item \cite{li_arxiv16}:
This work modifies BS by introducing an intra-sibling rank. For each partial solution, the set of $|\calV|$ continuations are sorted and assigned intra-sibling ranks $k\in[L]$ in order of decreasing log-probabilities, $\theta_t(y_t)$. 
The log-probability of an extenstion is then reduced in proportion to its rank, and continuations are re-sorted under these modified log-probabilities to select the top-B \emph{diverse} beam extensions. 
\item \cite{li_arxiv15}:
These models are decoded using a modified objective, $P(\mathbf{y}|x) - \lambda U(\mathbf{y})$, where $U(\mathbf{y})$ is an unconditioned target sequence model. 
This additional term penalizes generic input independent decoding.
\end{compactenum}
Both works use secondary mechanisms such as \emph{re-rankers} to pick a single solution from the generated lists. 
As we are interested in evaluating the quality of the generated lists and in isolating the gains due to diverse decoding, we do not implement any re-rankers. Instead, we simply sort the list based on log-probability.
We compare to our own implementations of these methods as none are publicly available.

\xhdr{Evaluation Metrics.} We evaluate the performance of the generated lists using the following two metrics that quantify complementary details:
\begin{compactenum}[\hspace{0pt} -]
\item \emph{Oracle Accuracy}:
Oracle or top-$k$ accuracy for some task-specific metric like BLEU is the maximum value of the metric over a list of $k$ potential solutions.
It is an upper bound on the potential impact diversity plays in finding relevant solutions. \\
\item \emph{Diversity Statistics}:
We count the number of distinct n-grams present in the list of generated outputs.
Similar to \cite{li_arxiv15}, we divide these counts by the total number of words generated to bias against long sentences.
\end{compactenum}
\emph{Simultaneous improvements} in both metrics indicate that output lists have increased diversity without sacrificing fluency and correctness with respect to target tasks.
Human preference studies which compare image captions produced by DBS and BS also compare these methods. 
Finally, We discuss the role of diversity by relating it to intrinsic details contained in images. 
\vspace{-10pt}
\subsection{Sensitivity Analysis and Effect of Diversity Functions} \label{div_choice}
\vspace{-5pt}
\label{div_type}
In this section, we study the impact of the number of groups, the strength of diversity penalty, and various forms of diversity functions for language models. 
Further discussion and experimental details are included in the supplementary materials.

\textbf{Number of Groups ($\mathbf{G}$).} Setting $G{=}B$ allows for the maximum exploration of the space, while setting $G{=}1$ reduces our method to BS, resulting in increased exploitation of the search-space around the 1-best decoding.
Thus, increasing the number of groups enables us to explore various modes of the model.
Empirically, we find that maximum exploration correlates with improved oracle accuracy and hence use $G{=}B$ to report results unless mentioned otherwise.

\textbf{Diversity Strength ($\mathbf{\lambda}$).} The diversity strength $\lambda$ specifies the trade-off between the joint $\log$-probability and the diversity terms. 
As expected, we find that a higher value of $\lambda$ produces a more diverse list; however, excessively high values of $\lambda$ can overpower model probability and result in grammatically incorrect outputs. 
We set $\lambda$ by performing a grid search on the validation set for all experiments.
We find a wide range of $\lambda$ values (0.2 to 0.8) work well for most tasks and datasets.

\textbf{Choice of Diversity Function ($\delta$).} 
\ar{As mentioned in \ref{form}, the sequence level dissimilarity term $\delta(\cdot, \cdot)$ can be design to satisfy different design choices.}
We discuss some of these below: 
\setlength{\plitemsep}{-1.5ex}
\begin{compactenum}[\hspace{0pt}-]
\item \emph{Hamming Diversity.} 
This form penalizes the selection of tokens used in previous groups proportional to the number of times it was selected before. \\
\item \emph{Cumulative Diversity.}
Once two sequences have diverged sufficiently, it seems unnecessary and perhaps harmful to restrict that they cannot use the same words at the same time. 
To encode this `backing-off' of the diversity penalty we introduce cumulative diversity which keeps a count of identical words used at every time step, indicative of overall dissimilarity.
Specifically, $\delta(\yb_{[t]}, \yb_{b,[t]}^g) = \exp\{-\nicefrac{\left(\sum_{\tau{\in}t}\sum_{b{\in}B'}I[y_{b,\tau}^h\neq y_{b,\tau}^g]\right)}{\Gamma}\}$ where $\Gamma$ is a temperature parameter controlling the strength of the cumulative diversity term and $I[\cdot]$ is the indicator function. \\
\item \emph{n-gram Diversity.}
The current group is penalized for producing the same n-grams as previous groups, regardless of alignment in time -- similar to \cite{gimpel_emnlp13}.
This is proportional to the number of times each n-gram in a candidate occurred in previous groups.
Unlike hamming diversity, n-grams capture higher order structures in the sequences. \\
\item \emph{Neural-embedding Diversity.}
While all the previous diversity functions discussed above perform exact matches, neural embeddings such as word2vec \citep{mikolov_nips13} can penalize semantically similar words like synonyms.
This is incorporated in each of the previous diversity functions by replacing the hamming similarity with a soft version obtained by computing the cosine similarity between word2vec representations.  
When using with n-gram diversity, the representation of the n-gram is obtained by summing the vectors of the constituent words. 
\end{compactenum}
Each of these various forms encode different notions of diversity.
Hamming diversity ensures different words are used at different times, but can be circumvented by small changes in sequence alignment. 
While n-gram diversity captures higher order statistics, it ignores sentence alignment. 
Neural-embedding based encodings can be seen as a semantic blurring of either the hamming or n-gram metrics, with word2vec representation similarity propagating diversity penalties not only to exact matches but also to close synonyms. 
We find that using any of the above functions help outperform BS in the tasks we examine; hamming diversity achieves the best oracle performance despite its simplicity. 
A comparison of the performance of these functions for image-captioning is provided in the supplementary.

\vspace{-5pt}

\subsection{Estimating Image Complexity}
Diversity in the output space is often dependent on the input. For example, ``complex'' scenes consisting of various objects and interactions tend to be described in multiple ways as compared to ``simple'' images that tend to focus on one specific object.
We study this by inspecting the gains due to DBS with varying complexity of images.
One notion of image complexity is studied by Ionescu \etal \cite{ionescu_cvpr16}, defining a ``difficulty score'' as the human response time for solving a visual search task for images in PASCAL-50S \cite{vedantam_cvpr15}.
Using the data from \cite{ionescu_cvpr16}, we train a Support Vector Regressor on ResNet \citep{he2016deep} features to predict this difficulty score. This model achieves a 0.41 correlation with the ground truth (comparable to the best model of \cite{ionescu_cvpr16} at 0.47). 
To evaluate the relationship between image complexity and performance gains from diverse decoding, we use this trained predictor to estimate a difficulty score $s$ for each image in the COCO \cite{coco} dataset. We compute the mean ($\mu=3.3$) and standard deviation ($\sigma=0.61$) and divide the images into three bins, \texttt{Simple} ($s\leq \mu-\sigma$), \texttt{Average} ($\mu{-}\sigma > s <\mu{+}\sigma$), and \texttt{Complex} ($s \geq \mu+\sigma$) consisting of 745, 3416 and 839 images respectively. 
Figure 3 shows some sample \texttt{Simple}, \texttt{Average}, and \texttt{Complex} images from the PASCAL-50S dataset. While simple images like close-up pictures of cats may only be described in a handful of ways by human captioners (first column), complex images with multiple objects and interactions will be described in many different ways depending on what is the focus of the captioner (last column). 
In the subsequent experiments on image-grounded language generation tasks, we show that improvements from DBS are greater for more complex images.

\subsection{Image Captioning}
\vspace{-5pt}
\xhdr{Dataset and Models.} We evaluate on two datasets -- COCO \citep{coco} and PASCAL-50S \citep{vedantam_cvpr15}. 
We use the public splits as in \cite{karpathy_cvpr15} for COCO.
PASCAL-50S is used only for testing save 200 validation images used to tune hyperparameters.
We train a captioning model \citep{vinyals_cvpr15} using the \texttt{neuraltalk2}\footnote{\texttt{\url{https://github.com/karpathy/neuraltalk2}}} code repository. 

\xhdr{Results.} As it can be observed from \tabref{tab:coco_quant} \ar{(Top)}, DBS outperforms both BS and \cite{li_arxiv16} on both COCO and PASCAL-50S datasets.
We observe that gains on PASCAL-50S are more pronounced (7.24\% and 9.60\% Oracle@20 improvements against BS and \cite{li_arxiv16}) than COCO.
This suggests diverse predictions are especially advantageous when there is a mismatch between training and testing sets making DBS a better inference strategy in real-world applications.

Table \ref{tab:coco_quant} \ar{(Top)} also shows the number of distinct n-grams produced by different techniques. 
Our method produces significantly more distinct n-grams (almost 300\% increase in the number of 4-grams produced) as compared to BS. 
We also note that our method tends to produce slightly longer captions compared to beam search on average.
Moreover, on the PASCAL-50S test split we observe that DBS finds more likely top-1 solutions on average -- DBS obtains a maximum $\log$-probability of -6.53 as against -6.91 got by BS of same beam width.
While the performance of DBS is guaranteed to be better than a BS of size $\nicefrac{B}{G}$, this experimental evidence suggests that using DBS as a replacement to BS leads to better or at least comparable performance.

\xhdr{Results by Image Complexity.}
From Table \ref{tab:coco_quant}, we can see that as the complexity of images increases DBS outperforms standard beam search (difference shown in parentheses) and other baselines by larger margins for all values of $k$.
For example, at Oracle Spice@20, DBS achieves significant improvements over BS of 0.67, 0.91, and 1.13 for \texttt{Simple}, \texttt{Average}, and \texttt{Complex} images respectively.
While DBS improves over BS in all settings, complex images benefit even more from diversity-inducing inference than simple images.
\newcommand{\imin}[1]{\includegraphics[width=60px, height=40px]{#1}}
\newcommand{\iminb}[1]{\setlength{\fboxsep}{-2pt}\setlength{\fboxrule}{2pt}\fbox{\includegraphics[width=60px, height=40px]{#1}}}
\begin{figure*}
    \caption{\ar{\textbf{A)} Sample PASCAL-50S images of different difficulty. Simple images are often close-ups of single objects while complex images involve  multiple objects in a wider view. \textbf{B)} Random human captions for the black-bordered images. Complex images have more varied captions than simpler images. \textbf{C}) which are not captured well by beam search compared to \textbf{D)} DBS.}}
    \vspace{5pt}
    \centering
    \setlength{\tabcolsep}{2.3pt}
    \renewcommand*{\arraystretch}{0.9}
    \resizebox{0.95\textwidth}{!}{
        \begin{tabular}{c | c c c | c c c | c c c}
            \multirow{3}{*}{\rotatebox[origin=c]{90}{\small \textbf{A) Sample Images~~~~~~~~}}~} & \multicolumn{3}{|c}{\textbf{\texttt{Simple}}} & \multicolumn{3}{| c |}{\textbf{\texttt{Average}}} & \multicolumn{3}{c}{\textbf{\texttt{Complex}}}\\[2pt]
             &  ~~~\imin{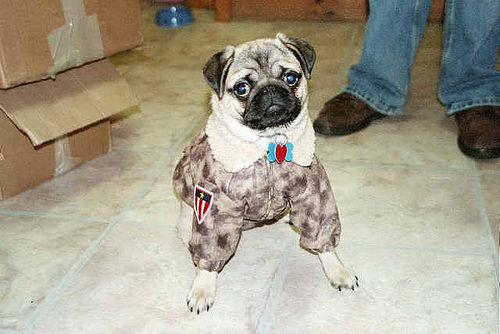} 
            & \imin{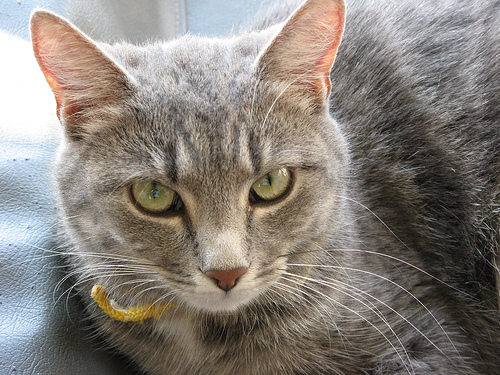} 
            & \imin{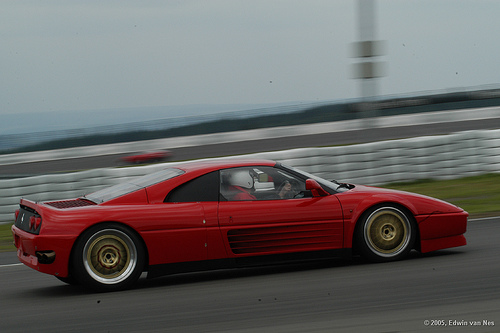}~~~~
              & ~~~~\imin{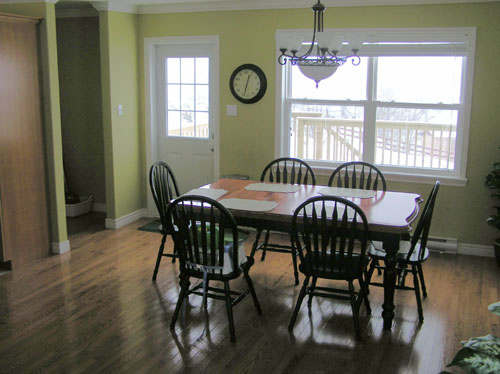} 
              & \imin{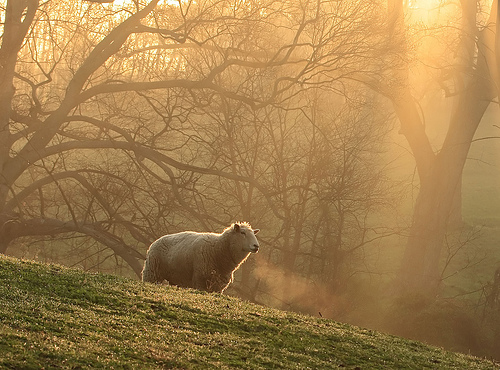}
              & \imin{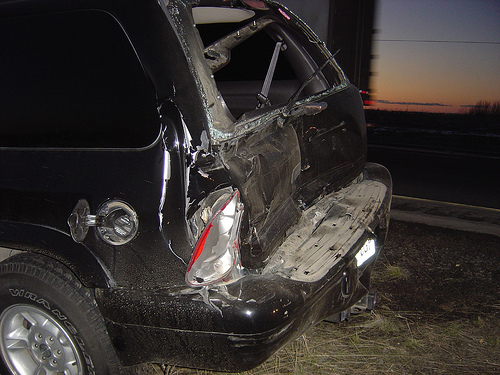}~~~~
                & ~~~~\imin{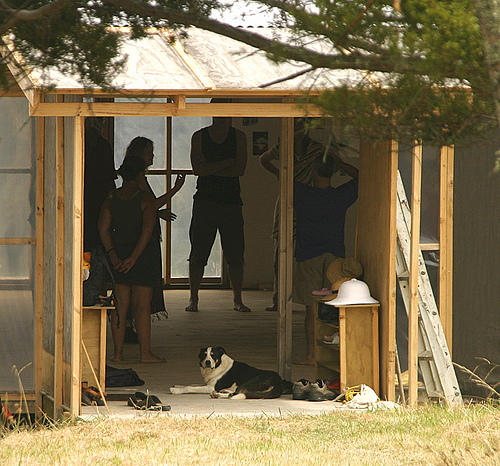} 
                & \imin{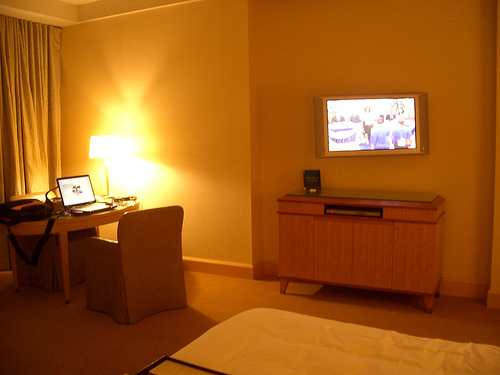}
                & \imin{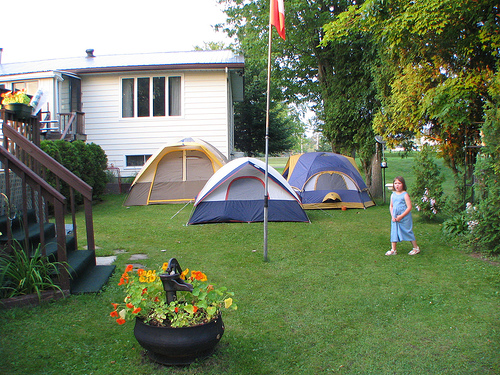}\\[-1pt]
              
            &  ~~~\imin{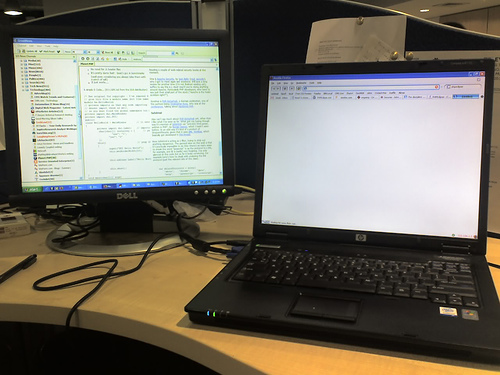} 
            & \iminb{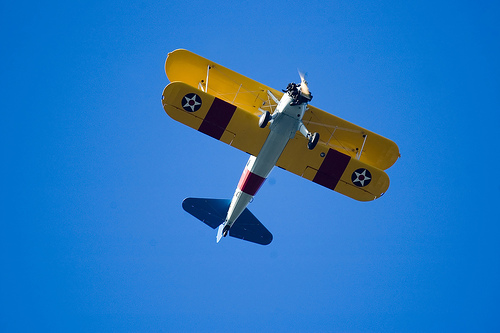} 
            & \imin{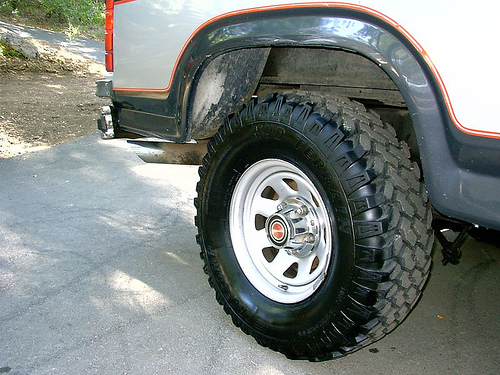}~~~~
              & ~~~~\imin{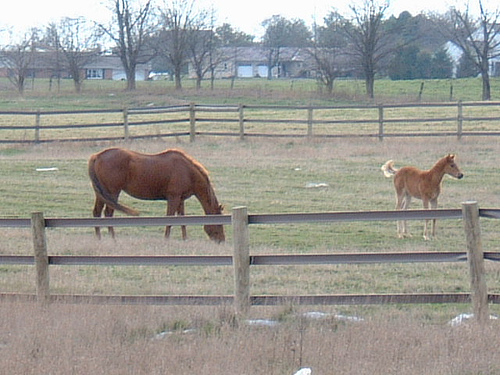} 
              & \iminb{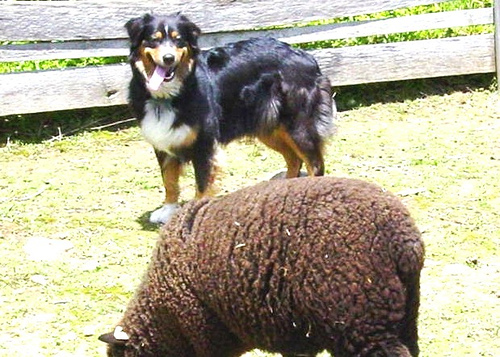}
              & \imin{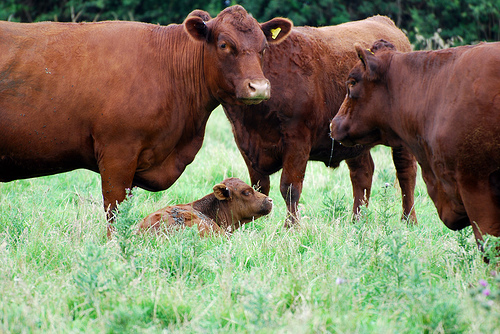}~~~~
                & ~~~~\imin{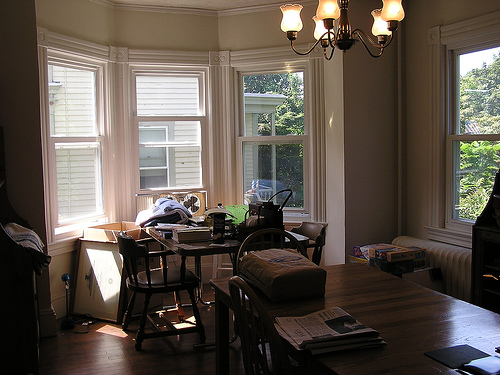} 
                & \iminb{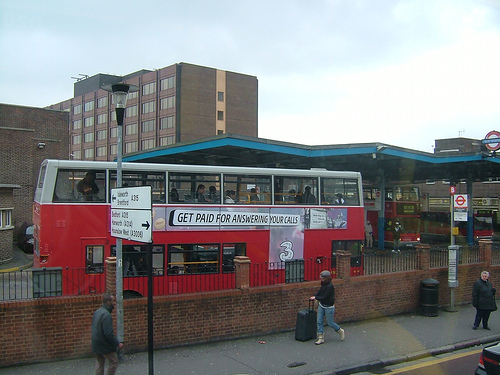}
                & \imin{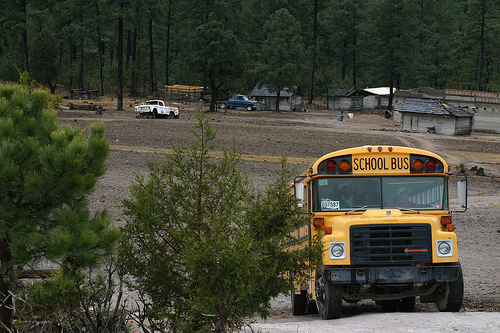}\\
             & \multicolumn{3}{c}{} & \multicolumn{3}{ c }{} & \multicolumn{3}{c}{}\\[-8pt]
            \cmidrule(l{30pt}r{30pt}){2-4}  \cmidrule(l{30pt}r{30pt}){5-7}  \cmidrule(l{30pt}r{30pt}){8-10} \\[-8pt]
            \multirow{4}{*}{\rotatebox[origin=c]{90}{\small \textbf{B) Human}}~}&\multicolumn{3}{| l}{~~~~\small A propeller plane is flying overhead} & \multicolumn{3}{ l }{~~~~\small A black sheep dog watches over a black sheep.} & \multicolumn{3}{l}{~~~\small A double-decker bus is pulling into a bus station.}\\
            &\multicolumn{3}{| l}{~~~~\small A old time airplane perform in the air show.} & \multicolumn{3}{ l }{~~~~\small A dog and lamb are playing in a fenced area.} & \multicolumn{3}{l}{~~~\small People walking past a red and white colored bus.}\\
            &\multicolumn{3}{|l}{~~~~\small A small plane is flying through the air.} & \multicolumn{3}{ l }{~~~~\small A black dog looking at a brown sheep in a field.} & \multicolumn{3}{l}{~~~\small A double-decker bus pulls into a terminal.}\\
            &\multicolumn{3}{|l}{~~~~\small The biplane with the yellow wings flew in the sky.} & \multicolumn{3}{ l }{~~~~\small A dog is standing near a sheep.} & \multicolumn{3}{l}{~~~\small People walk down the sidewalk at a bus station.}\\
            &  \multicolumn{3}{c}{} & \multicolumn{3}{ c }{} & \multicolumn{3}{c}{}\\[-8pt]
            \cmidrule(l{30pt}r{30pt}){2-4}  \cmidrule(l{30pt}r{30pt}){5-7}  \cmidrule(l{30pt}r{30pt}){8-10} \\[-8pt]
            \multirow{4}{*}{\rotatebox[origin=c]{90}{\small \textbf{C) BS}}~}&\multicolumn{3}{| l}{~~~~\small A blue and yellow biplane flying in the sky.} & \multicolumn{3}{ l }{~~~~\small A dog sitting on the ground next to a fence.} & \multicolumn{3}{l}{~~~\small A red double decker bus driving down a street.}\\
            &\multicolumn{3}{| l}{~~~~\small A small airplane is flying in the sky.} & \multicolumn{3}{ l }{~~~~\small A black and white dog standing next to a sheep.} & \multicolumn{3}{l}{~~~\small A double decker bus parked in front of a building.}\\
            &\multicolumn{3}{| l}{~~~~\small A blue and yellow biplane flying in the sky.} & \multicolumn{3}{ l }{~~~~\small A dog is sitting on the ground next to a fence.} & \multicolumn{3}{l}{~~~\small A double decker bus driving down a city street.}\\
            &\multicolumn{3}{| l}{~~~~\small A small airplane flying in the blue sky.} & \multicolumn{3}{ l }{~~~\small A black and white dog standing next to a dog.} & \multicolumn{3}{l}{~~~\small A double decker bus is parked on the side of the street.}\\
            &  \multicolumn{3}{c}{} & \multicolumn{3}{ c }{} & \multicolumn{3}{c}{}\\[-8pt]
            \cmidrule(l{30pt}r{30pt}){2-4}  \cmidrule(l{30pt}r{30pt}){5-7}  \cmidrule(l{30pt}r{30pt}){8-10} \\[-8pt]
            \multirow{4}{*}{\rotatebox[origin=c]{90}{\small \textbf{D) DBS}}~}&\multicolumn{3}{| l}{~~~~\small A small airplane flying through a blue sky.} & \multicolumn{3}{ l }{~~~~\small There is a dog that is sitting on the ground.} & \multicolumn{3}{l}{~~~~\small A red double decker bus driving down a street.}\\
            &\multicolumn{3}{| l}{~~~~\small A blue and yellow biplane flying in the sky.} & \multicolumn{3}{ l }{~~~~\small An animal that is laying down in the grass.} & \multicolumn{3}{l}{~~~\small The city bus is traveling down the street.}\\
            &\multicolumn{3}{| l}{~~~~\small There is a small plane flying in the sky.} & \multicolumn{3}{ l }{~~~~\small There is a black and white dog sitting on the ground.} & \multicolumn{3}{l}{~~~\small People are standing in front of a double decker bus.}\\
            &\multicolumn{3}{| l}{~~~~\small An airplane flying with a blue sky in the background.} & \multicolumn{3}{ l }{~~~~\small Two dogs are sitting on the ground with a fence.} & \multicolumn{3}{l}{~~~\small The city bus is parked on the side of the street.}\\
    \end{tabular}}
    \label{fig:diff}
\end{figure*}

\ar{\xhdr{Human Preference by Difficulty.} To further establish the effectiveness of our method, we evaluate human preference between captions decoded using DBS and BS.
In this forced-choice test, DBS captions were preferred over BS 60\% of the time by human annotators.
Further, they were preferred about 50\%, 69\% and 83\% of the times  for \texttt{Simple}, \texttt{Average} and \texttt{Difficult} images  respectively.
Furthermore, we observe a positive correlation ($\rho = 0.73$) between difficulty scores and humans preferring DBS to BS.
Further details about this experiment are provided in the supplement.}
\begin{table}[t]
    \caption{\ar{\textbf{Top:} Oracle SPICE@$k$ and distinct n-grams on the COCO image captioning task at $B=20$. While we report SPICE, we observe similar trends in other metrics (reported in the supplement). \textbf{Bottom:} Breakdown of results by difficulty class, highlighting the relative improvement over BS.}}
\centering
\renewcommand*{\arraystretch}{1.1}
\setlength{\tabcolsep}{5pt}
\resizebox{0.7\columnwidth}{!}{
    \begin{tabular}{ c c  c c c c  c c c c @{}}\toprule
         &\multirow{2}{*}{\textbf{Method}} &\multirow{2}{*}{\textbf{SPICE}} &  \multicolumn{3}{c}{\textbf{Oracle SPICE@k}} & \multicolumn{4}{c}{\textbf{Distinct n-Grams}} \\
        \cmidrule[0.75pt](lr){4-6} \cmidrule[0.75pt](lr){7-10} 
          & &  & @5 & @10 & @20 & n = 1 & 2 & 3 & 4 \\
        \midrule
          \multirow{4}{*}{\rotatebox[origin=c]{90}{COCO}}& BS & 16.27 & 22.96 & 25.14 & 27.34 & 0.40 & 1.51 & 3.25 & 5.67  \\
          &\citet{li_arxiv16} & 16.35 & 22.71 & 25.23 & 27.59 & 0.54 & 2.40 & 5.69 & 8.94 \\
          & DBS & \textbf{16.783} & \textbf{23.08} & \textbf{26.08} & \textbf{28.09} & \textbf{0.56} & \textbf{2.96} & \textbf{7.38} & \textbf{13.44} \\
        \cmidrule[0.75pt](lr){2-10}
        & \citet{li_arxiv15} & 16.74 & 23.27 & 26.10 & 27.94 & 0.42 & 1.37 & 3.46 & 6.10 \\
        \midrule
          \multirow{4}{*}{\rotatebox[origin=c]{90}{PASCAL-50S}} & BS & 4.93 & 7.04 & 7.94 & 8.74 & 0.12 & 0.57 & 1.35 & 2.50  \\
          &\citet{li_arxiv16} & 5.08 & 7.24 & 8.09 & 8.91 & 0.15 & 0.97 & 2.43 & 5.31 \\
          &DBS & \textbf{5.357} & \textbf{7.357} & \textbf{8.269} & \textbf{9.293} & \textbf{0.18} & \textbf{1.26} & \textbf{3.67} & \textbf{7.33} \\
        \cmidrule[0.75pt](lr){2-10}
         &\citet{li_arxiv15} & 5.12 & 7.17 & 8.16 & 8.56 & 0.13 & 1.15 & 3.58 & 8.42 \\
\end{tabular}}\\[5pt]
\renewcommand*{\arraystretch}{1.2}
\setlength{\tabcolsep}{5pt}
\resizebox{0.7\columnwidth}{!}{
    \begin{tabular}{ c c  c c c c }\toprule
         &\multirow{2}{*}{\textbf{Method}} & \multirow{2}{*}{\textbf{SPICE}} & \multicolumn{3}{c}{\textbf{Oracle SPICE@$k$ (Gain over BS)}} \\
        \cmidrule[0.75pt](lr){4-6} 
          & & & @5 & @10 & @20 \\
        \midrule
          \multirow{4}{*}{\rotatebox[origin=c]{90}{Simple}}
            & BS & 17.28 (0) & 24.32 (0) & 26.73 (0) & 28.7 (0)  \\
            &\citet{li_arxiv16} & 17.12 (-0.16) & 24.17 (-0.15) & 26.64 (-0.09) & 29.28 (0.58)\\
            & DBS & \textbf{17.42 (0.14)} & \textbf{24.44 (0.12)} & \textbf{26.92 (0.19)} & \textbf{29.37 (0.67)} \\
          \cmidrule[0.75pt](lr){2-6}
          &\citet{li_arxiv15} & 17.38 (0.1) & 24.48 (0.16) & 26.82 (0.09) & 29.21 (0.51) \\
          \midrule
            \multirow{4}{*}{\rotatebox[origin=c]{90}{Average}}
              & BS & 15.95 (0) & 22.51 (0) & 24.8 (0) & 26.55 (0)  \\
              &\citet{li_arxiv16} & 16.19 (0.24) & 22.59 (0.08) & 24.98 (0.18) & 27.23 (0.68)\\
              & DBS & \textbf{16.28 (0.33)} & \textbf{22.65 (0.14)} & \textbf{25.08 (0.28)} & \textbf{27.46 (0.91)} \\
            \cmidrule[0.75pt](lr){2-6}
            &\citet{li_arxiv15} & 16.22 (0.27) & 22.61 (0.1) & 25.01 (0.21) & 27.12 (0.57) \\
            \midrule
              \multirow{4}{*}{\rotatebox[origin=c]{90}{Complex}}
                & BS & 16.39 (0) & 22.62 (0) & 24.91 (0) & 27.23 (0)  \\
                &\citet{li_arxiv16} & 16.55 (0.16) & 22.55 (-0.07) & 25.18 (0.27) & 27.57 (0.34)\\
                & DBS & \textbf{16.75 (0.36)} & \textbf{22.81 (0.19)} & \textbf{25.25 (0.34)} & \textbf{28.36 (1.13)} \\
              \cmidrule[0.75pt](lr){2-6}
              &\citet{li_arxiv15} & 16.69 (0.3) & 22.69 (0.07) & 25.16 (0.25) & 27.94 (0.71) \\
              \bottomrule
      \end{tabular}}
      \label{tab:coco_quant}
  \end{table} 

\begin{figure}[h]
\includegraphics[width=\columnwidth, clip=true, trim=0 70pt 0 30pt,height=5cm]{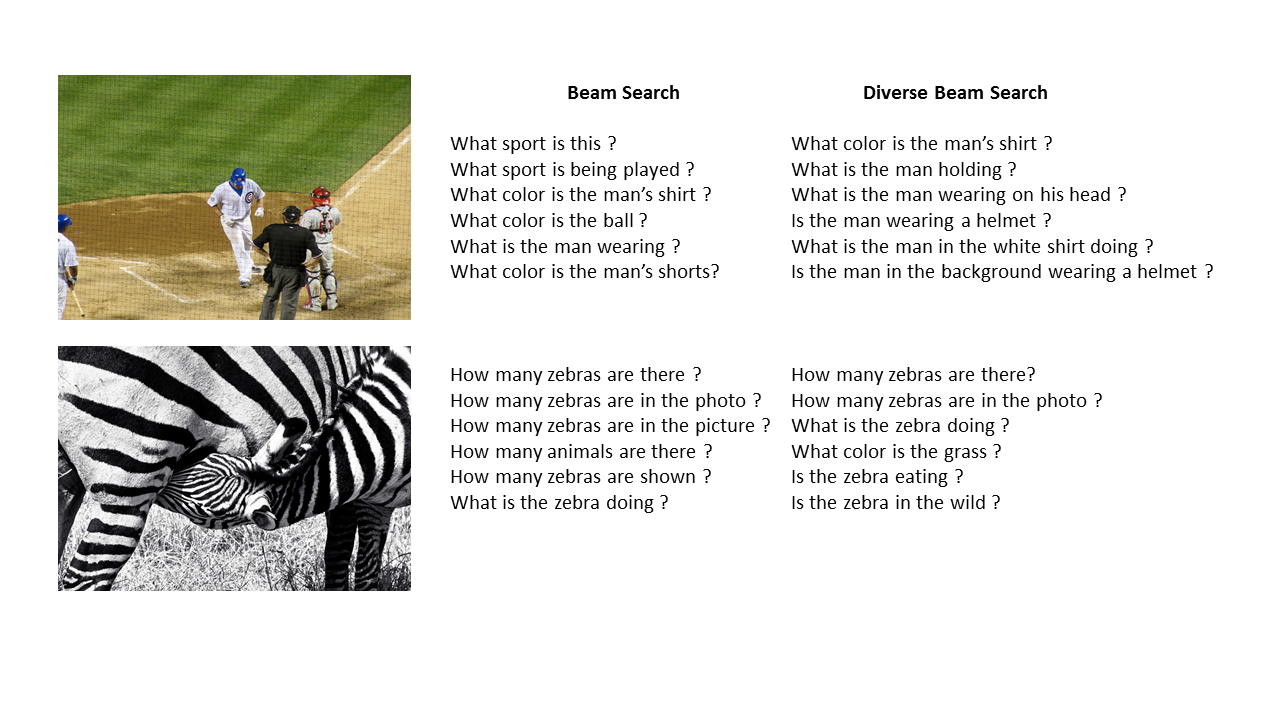}
\caption{Qualitative results on Visual Question Generation. It can be noted that DBS generates questions that are non-generic which belong to different question types}
\label{fig:vqg}
\vspace{-2ex}
\end{figure}



\subsection{Visual Question Generation}
We also report results on Visual Question Generation (VQG) \citep{mostafazadeh_arxiv16}, where a model is trained to produce questions \emph{about an image}.
Generating visually focused questions requires reasoning about multiple problems that are central to vision -- \eg, object attributes, relationships between objects, and natural language.
Similar to captioning, there are many sensible questions for a given image. 

The VQG dataset \citep{mostafazadeh_arxiv16} consists of 5 human-generated questions per image for 5000 images from COCO \citep{coco}.
We use a model similar to the one used for captioning, except that it is now trained to output questions rather than captions.
Similar to previous results, using beam search to sample outputs results in similarly worded question while DBS decoded questions ask about multiple details of the image (see Fig.~\ref{fig:vqg}). 

We show quantitative evaluations in Table \ref{tab:vqg_quant} for the VQG dataset as a whole and when partitioned by image difficulty. We find DBS significantly outperforms the baseline methods on this task both on standard metrics (SPICE) and measure of diversity. We also observe that gap between DBS and the baseline methods is more pronounced than in the captioning task and attribute this to the increased variety of possible visually grounded questions compared to captions which often describe only a few major salient objects. The general trend that more complex images benefit more from diverse decoding also persists in this setting.


\begin{table}
    \caption{\ar{\textbf{Top:} Oracle SPICE@$k$ and distinct n-grams on the VQG task at $B=20$. \textbf{Bottom:} Results by difficulty class, highlighting the relative improvement over BS.}}
    \centering
    \renewcommand*{\arraystretch}{1.1}
    \setlength{\tabcolsep}{5pt}
    \resizebox{0.7\columnwidth}{!}{
        \begin{tabular}{ c c  c c c c  c c c c @{}}\toprule
             &\multirow{2}{*}{\textbf{Method}} &\multirow{2}{*}{\textbf{SPICE}} &  \multicolumn{3}{c}{\textbf{Oracle SPICE@k}} & \multicolumn{4}{c}{\textbf{Distinct n-Grams}} \\
            \cmidrule[0.75pt](lr){4-6} \cmidrule[0.75pt](lr){7-10} 
              & & & @5 & @10 & @20 & n = 1 & 2 & 3 & 4 \\
            \midrule
              \multirow{4}{*}{\rotatebox[origin=c]{90}{VQG}}& BS & 15.17 & 21.96 & 23.16 & 26.74 & 0.31 & 1.36 & 3.15 & 5.23  \\
              &\citet{li_arxiv16} & 15.45 & 22.41 & 25.23 & 27.59 & 0.34 & 2.40 & 5.69 & 8.94 \\
              & DBS & \textbf{16.49} & \textbf{23.11} & \textbf{25.71} & \textbf{27.94} & \textbf{0.43} & \textbf{2.17} & \textbf{6.49} & \textbf{12.24} \\
            \cmidrule[0.75pt](lr){2-10}
            & \citet{li_arxiv15} & 16.34 & 22.92 & 25.12 & 27.19 & 0.35 & 1.56 & 3.69 & 7.21 \\
    \end{tabular}}\\[5pt]
    \renewcommand*{\arraystretch}{1.2}
    \setlength{\tabcolsep}{5pt}
    \resizebox{0.7\columnwidth}{!}{
        \begin{tabular}{ c c  c c c c }\toprule
             &\multirow{2}{*}{\textbf{Method}} &  \multirow{2}{*}{\textbf{SPICE}} & \multicolumn{3}{c}{\textbf{Oracle SPICE@$k$ (Gain over BS)}} \\
            \cmidrule[0.75pt](lr){4-6} 
              & & & @5 & @10 & @20 \\
            \midrule
              \multirow{4}{*}{\rotatebox[origin=c]{90}{Simple}}
                & BS & 16.04 (0) & 21.34 (0) & 23.98 (0) & 26.62 (0)  \\
                &\citet{li_arxiv16} & 16.12 (0.12) & 21.65 (0.31) & 24.64 (0.66) & 26.68 (0.04)\\
                & DBS & \textbf{16.42 (0.38)} & \textbf{22.44 (1.10)} & \textbf{24.71 (0.73)} & \textbf{26.73 (0.13)} \\
              \cmidrule[0.75pt](lr){2-6}
              &\citet{li_arxiv15} & 16.18 (0.14) & 22.18 (0.74) & 24.16 (0.18) & 26.23 (-0.39) \\
              \midrule
                \multirow{4}{*}{\rotatebox[origin=c]{90}{Average}}
                  & BS & 15.29 (0) & 21.61 (0) & 24.12 (0) & 26.55 (0)  \\
                  &\citet{li_arxiv16} & 16.20 (0.91) & 21.90 (0.29) & 25.61 (1.49) & 27.41 (0.86)\\
                  & DBS & \textbf{16.63 (1.34)} & \textbf{22.81 (1.20)} & \textbf{24.68 (0.46)} & \textbf{27.10 (0.55)} \\
                \cmidrule[0.75pt](lr){2-6}
                &\citet{li_arxiv15} & 16.07 (0.78) & 22.12 (-0.49) & 24.34 (0.22) & 26.98 (0.43) \\
                \midrule
                  \multirow{4}{*}{\rotatebox[origin=c]{90}{Complex}}
                    & BS & 15.78 (0) & 22.41 (0) & 24.48 (0) & 26.87 (0)  \\
                    &\citet{li_arxiv16} & 16.82 (1.04) & 23.20 (0.79) & 25.48 (1.00) & 27.12 (0.25)\\
                    & DBS & \textbf{17.25 (1.47)} & \textbf{23.35(1.13)} & \textbf{26.19 (1.71)} & \textbf{28.01 (1.03)} \\
                  \cmidrule[0.75pt](lr){2-6}
                  &\citet{li_arxiv15} & 17.10 (1.32) & 23.31 (0.90) & 26.01 (1.53) & 27.92 (1.05) \\
                  \bottomrule
          \end{tabular}}\\[5pt]
          \label{tab:vqg_quant}
      \end{table} 

\subsection{Machine Translation}
\xhdr{Dataset and Models.} We use the English-French parallel data from the \emph{europarl} corpus as the training set.
We report results on \emph{news-test-2013} and \emph{news-test-2014} and use the \emph{newstest2012} to tune DBS parameters.
We train a encoder-decoder architecture as proposed in \cite{bahdanau_arxiv14} using the \texttt{dl4mt-tutorial}\footnote{\url{https://github.com/nyu-dl/dl4mt-tutorial}} code repository. 
The encoder consists of a bi-directional recurrent network (Gated Recurrent Unit) with attention. 
We use sentence level BLEU scores \citep{papineni_acl02} to compute oracle metrics and report distinct n-grams similar to image-captioning. 
From \tabref{tab: mt_quant}, we see that DBS consistently outperforms standard baselines with respect to both metrics.

\begin{table}[h!]
\centering
\caption{Quantitative results on En-Fr machine translation on the newstest-2013 dataset (at $B=20$).
Although we report BLEU-4 values, we find similar trends hold for lower BLEU metrics as well.}
\resizebox{\columnwidth}{!}{
\begin{tabular}{c  c c c c  c c c c @{}}\toprule
\textbf{Method} & \multicolumn{4}{c}{\textbf{Oracle Accuracy (BLEU-4)}} & \multicolumn{4}{c}{\textbf{Diversity Statistics}} \\
\cmidrule[0.75pt](lr){2-5} \cmidrule[0.75pt](lr){6-9} 
 & @1 & @5 & @10 & @20 & distinct-1 & distinct-2 & distinct-3 & distinct-4 \\
\midrule
Beam Search & 13.52 & 16.67 & 17.63 & 18.44 & 0.04 & 0.75 & 2.10 & 3.23  \\
\cite{li_arxiv16} & 13.63 & 17.11 & 17.50 & 18.34 & 0.04 & 0.81 & 2.92 & 4.61 \\
DBS & \textbf{13.69} & \textbf{17.51} & \textbf{17.80} & \textbf{18.77} & \textbf{0.06} & \textbf{0.95} & \textbf{3.67} & \textbf{5.54}\\
\midrule
\cite{li_arxiv15} & 13.40 & 17.54 & 17.97 & 18.86 & 0.04 & 0.86 & 2.76 & 4.31\\
\bottomrule
\end{tabular}}
\label{tab: mt_quant}
\end{table} 


\section{Conclusion}
Beam search is the most commonly used approximate inference algorithm to decode sequences from RNNs; however, it suffers from a lack of diversity.
Producing multiple highly similar and generic outputs is not only wasteful in terms of computation but also detrimental for tasks with inherent ambiguity like image captioning.
In this work, we presented \emph{Diverse Beam Search}, which describes beam search as an optimization problem and augments the objective with a diversity term.
The result is a `doubly greedy' approximate algorithm that produces diverse decodings while using about the same time and resources as beam search.
Our method consistently outperforms beam search and other baselines across all our experiments without \emph{extra computation} or \emph{task-specific overhead}.
\ar{Further, in the case of image-grounded language generation tasks, we find that DBS provides increased gains as the complexity of the images increases.}
DBS is \emph{task-agnostic} and can be applied to any case where BS is used -- making it applicable in multiple domains.




\bibliography{strings,ashwinkv}
\bibliographystyle{iclr2017_conference}

\clearpage
\section*{Appendix}

\subsection*{Sensivity Studies}

\begin{figure}[h]
\includegraphics[width =\columnwidth, height = 2 cm]{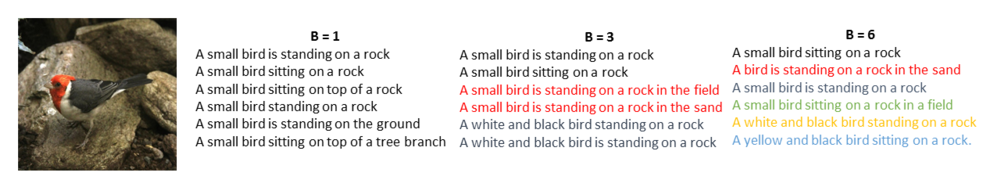}\\
\caption{Effect of increasing the number of groups $G$. The beams that belong to the same group are colored similarly. Recall that diversity is only enforced across groups such that $G=1$ corresponds to classical BS.}
\label{fig:G}
\end{figure}

\textbf{Number of Groups.} \figref{fig:G} presents snapshots of the transition from BS to DBS at $B=6$ and $G=\{1,3,6\}$. As beam width moves from 1 to $G$, the exploration of the method increases resulting in more diverse lists.

\begin{figure}[h]
    \begin{subfigure}[h]{0.5\textwidth}
	\centering    
    \includegraphics[width=\textwidth]{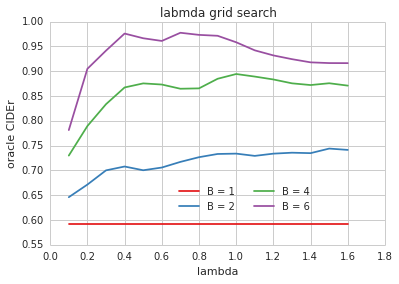}
    \caption{Grid search of diversity strength parameter}
    \label{fig:lambdagrid}
  \end{subfigure}
  \begin{subfigure}[h]{0.5\textwidth}
	\centering    
    \includegraphics[width=\textwidth]{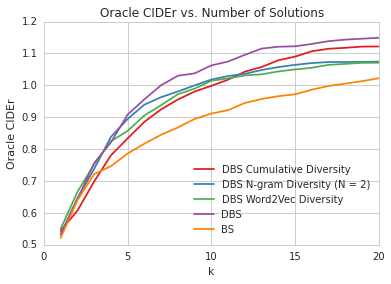}
    \caption{Effect of multiple forms for the diversity function}
    \label{fig:divtype}
  \end{subfigure}
  \vspace{10pt}
  \caption{
    \figref{fig:lambdagrid} shows the results of a grid search of the diversity strength ($\lambda$) parameter of DBS on the validation split of PASCAL 50S dataset. 
    We observe that it is robust for a wide range of values.
    \figref{fig:divtype} compares the performance of multiple forms for the diversity function ($\Delta$). 
While \naive diversity performs the best, other forms are comparable while being better than BS. }
\end{figure}

\textbf{Diversity Strength.} As noted in \secref{div_type}, our method is robust to a wide range of values of the diversity strength ($\lambda$). 
\figref{fig:lambdagrid} shows a grid search of $\lambda$ for image-captioning on the PASCAL-50S dataset. 

\textbf{Choice of Diversity Function.} \figref{fig:divtype} shows the oracle performace of various forms of the diversity function described in \secref{div_type}. We observe that hamming diversity surprisingly performs the best. Other forms perform comparably while outperforming BS.

\subsection*{Human Studies}
\begin{figure}[h]
    \centering
    \includegraphics[width=\columnwidth]{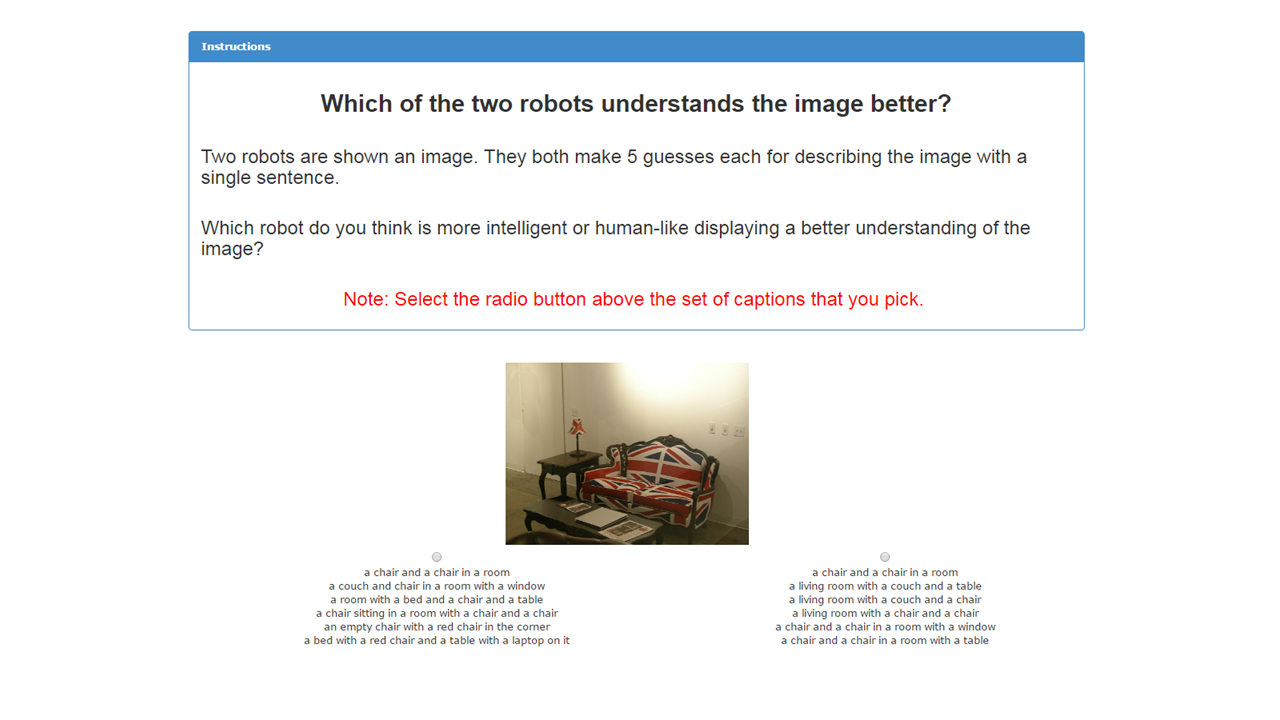}
    \caption{Screen-shot of the interface used to perform human studies}
    \label{fig:amt}
\end{figure}

For image-captioning, we conduct a human preference study between BS and DBS captions as explained in \secref{sec:results}. 
A screen shot of the interface used to collect human preferences for captions generated using DBS and BS is presented in \figref{fig:amt}.
The lists were shuffled to guard the task from being gamed by a turker.

\begin{table}[h!]
    \centering
    \begin{tabular}{ c c c }
        \toprule
        difficulty score  &  \# images & \% images DBS \\
        bin range         &            & was preffered \\
        \midrule
         $\leq \mu-\sigma$ & 481 & 50.51\% \\
        $[\mu{-}\sigma,\mu{+}\sigma]$  & 409 & 69.92\% \\
         $\geq \mu+\sigma$ & 110 & 83.63\%  \\
        \bottomrule
    \end{tabular}
    \vspace{10pt}
    \caption{Frequency table for image difficulty and human preference for DBS captions on PASCAL50S dataset}
    \label{tab:amt}
\end{table}

\begin{figure}[h]
    \centering
    \includegraphics[width=\columnwidth]{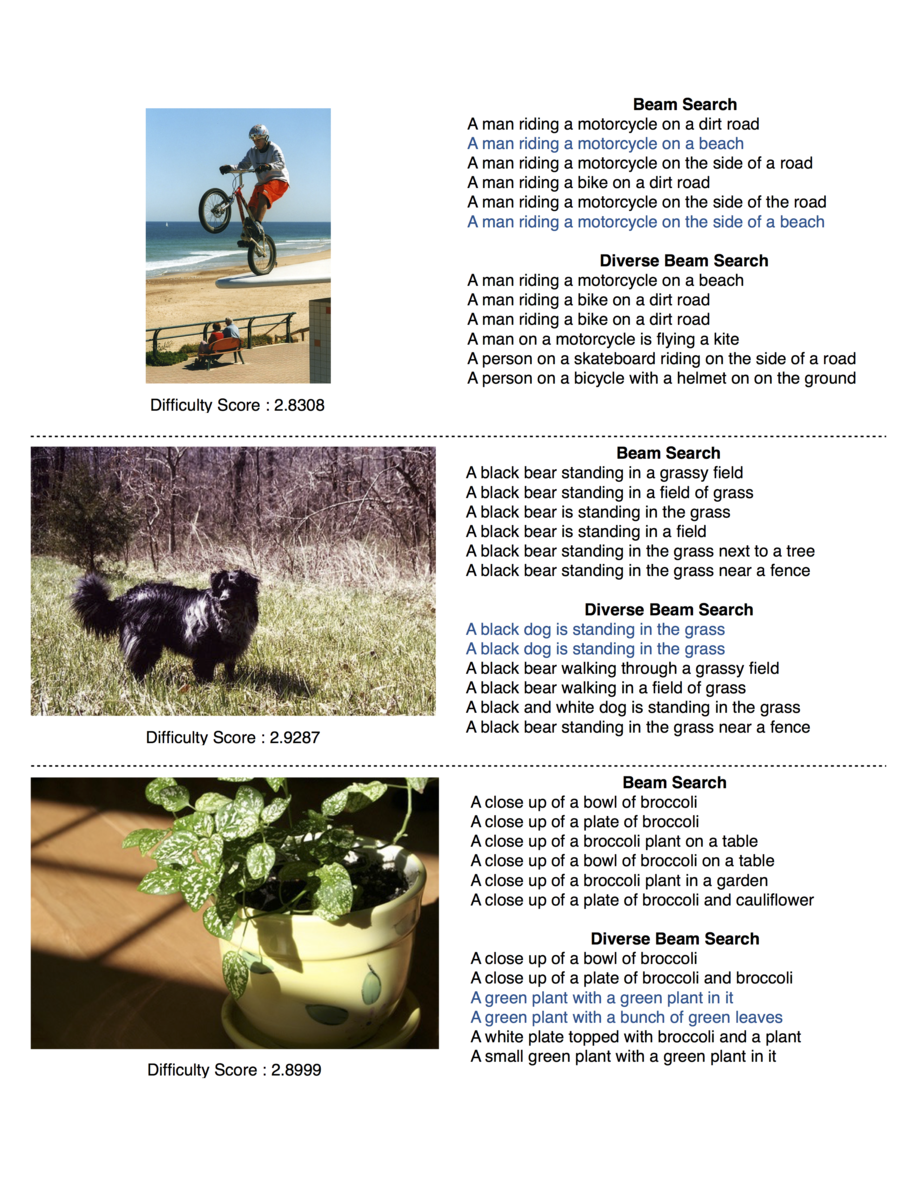}
\caption{For images with low difficulty score, BS captions are preferred to DBS -- as show in the first figure. However, we observe that DBS captions perform better when there is a mismatch between the statistics of the testing and training sets. Interesting captions are colored in blue for readability.}
    \label{fig:qual_2}
\end{figure}

\begin{figure}[h]
    \centering
    \includegraphics[width=\columnwidth]{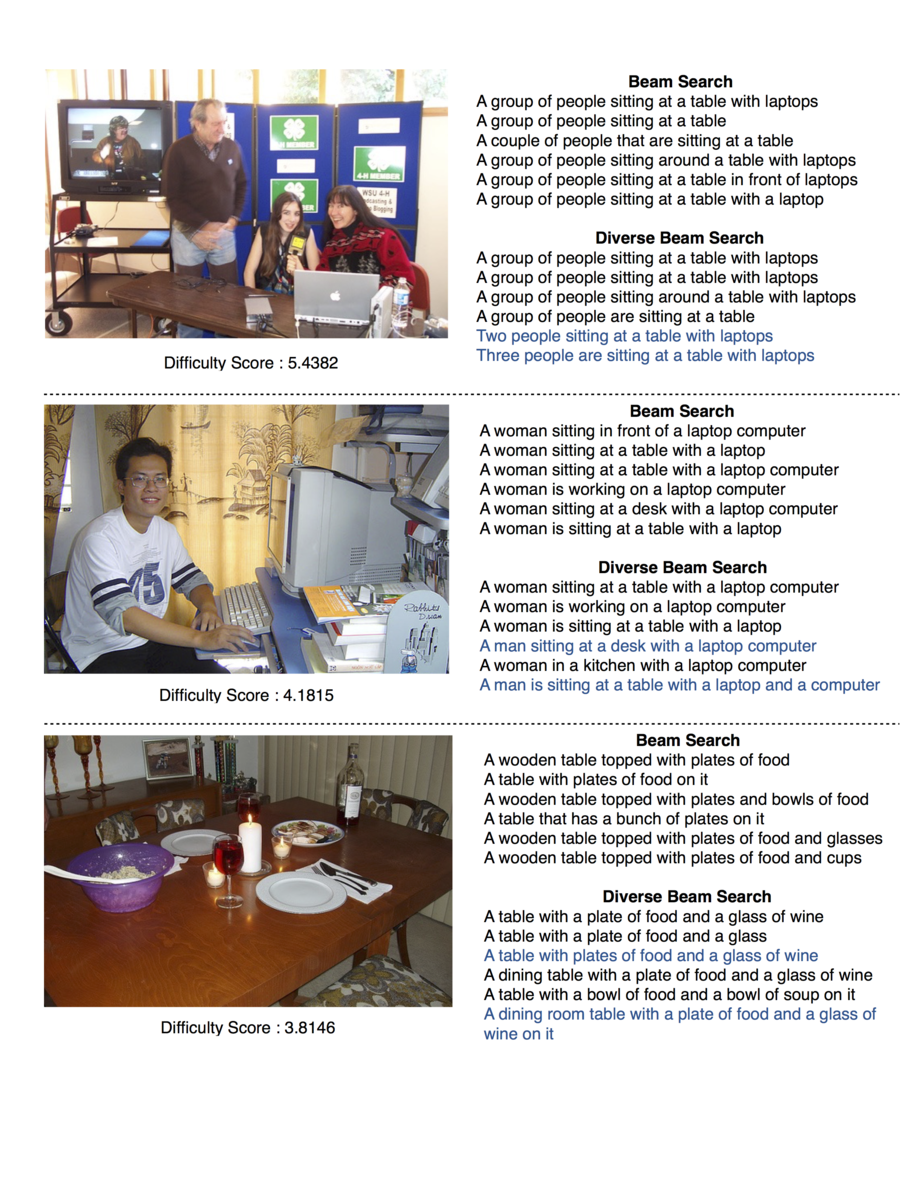}
    \caption{For images with a high difficulty score, captions produced by DBS are preferred to BS. 
    Interesting captions are colored in blue for readability. }
    \label{fig:qual_1}
\end{figure}

As mentioned in \secref{sec:results}, we observe that \emph{difficulty score} of an image and human preference for DBS captions are positively correlated. 
The dataset contains more images that are less difficulty and so, we analyze the correlation by dividing the data into three bins. 
For each bin, we report the \% of images for which DBS captions were preferred after a majority vote (\ie at least 3/5 turkers voted in favor of DBS) in \tabref{tab:amt}.
At low difficulty scores consisting mostly of iconic images -- one might expect that BS would be preferred more often than chance. 
However, mismatch between the statistics of the training and testing data results in a better performance of DBS.
Some examples for this case are provided in \figref{fig:qual_2}. 
More general qualitative examples are provided in \figref{fig:qual_1}.

\end{document}